\documentclass[journal]{IEEEtran}
\usepackage{graphicx,epsfig,amsfonts,amsmath,amssymb,url,ifthen}
\usepackage{times}
\usepackage{bm}
\usepackage{multirow}
\usepackage{subfigure}
\usepackage{algorithm}
\usepackage{algorithmic}

\usepackage{amsopn}
\usepackage{cases}
\usepackage{url}
\usepackage{epsfig}
\usepackage{graphicx,subfig}
\usepackage{graphicx}
\usepackage{subfigure}
\usepackage{cite}
\usepackage{amsthm}
\usepackage{CJK}
\usepackage{color}
\usepackage{makecell}
\usepackage{enumitem}
\usepackage{multirow,tabularx}
\graphicspath{{./Figures/}}

\newcommand{\tx}{\bm{x}}
\newcommand{\ty}{\bm{y}}

\newcommand{\te}{\bm{e}}

\newcommand{\tD}{\textbf{D}}

\newcommand{\tI}{\textbf{I}}

\newcommand{\tW}{\textbf{W}}

\newcommand{\tS}{\textbf{S}}
\newcommand{\tLambda}{\mathbf{\Lambda}}

\newcommand{\tw}{\bm{w}}

\newcommand{\tPhi}{\mathbf{\Phi}}

\newcommand{\valpha}{\bm{\alpha}}

\newcommand{\vphi}{\bm{\phi}}

\newcommand{\vTheta}{\bm{\Theta}}
\DeclareMathOperator*{\argmin}{argmin}

\title{ConvCSNet: A Convolutional Compressive Sensing Framework Based on Deep Learning}

\author{ Xiaotong Lu, Weisheng~Dong, ~\IEEEmembership{Member,~IEEE}, Peiyao Wang, Guangming Shi, ~\IEEEmembership{Senior member,~IEEE}, and Xuemei Xie   
\thanks{X. Lu, W. Dong, P. Wang, G. Shi, and X. Xie are with School of Electronic Engineering, Xidian University, Xi'an, 710071, China (e-mail: wsdong@mail.xidian.edu.cn)}
}

\newboolean{doublecolumn}
\setboolean{doublecolumn}{true}

\begin{document}
\maketitle

\ifthenelse {\boolean{doublecolumn}} {} {\baselineskip=.8cm}

\begin{abstract}
  Compressive sensing (CS), aiming to reconstruct an image/signal from a small set of random measurements has attracted considerable attentions in recent years. Due to the high dimensionality of images, previous CS methods mainly work on image blocks to avoid the huge requirements of memory and computation, i.e., image blocks are measured with Gaussian random matrices, and the whole images are recovered from the reconstructed image blocks. Though efficient, such methods suffer from serious blocking artifacts. In this paper, we propose a convolutional CS framework that senses the whole image using a set of convolutional filters. Instead of reconstructing individual blocks, the whole image is reconstructed from the linear convolutional measurements. Specifically, the convolutional CS is implemented based on a convolutional neural network (CNN), which performs both the convolutional CS and nonlinear reconstruction. Through end-to-end training, the sensing filters and the reconstruction network can be jointly optimized. To facilitate the design of the CS reconstruction network, a novel two-branch CNN inspired from a sparsity-based CS reconstruction model is developed. Experimental results show that the proposed method substantially outperforms previous state-of-the-art CS methods in term of both PSNR and visual quality.
\end{abstract}


\IEEEpeerreviewmaketitle

\section{Introduction}

Compressive sensing (CS) is a new image/signal acquisition framework acquiring only a few linear measurements \cite{Candes:IT:2006,Donoho:IT:2006} with random measurement matrices. Such framework has the potentials of significantly improving the imaging speed and sensor energy efficiency in real applications. Based on CS, several new imaging systems, including single-pixel camera \cite{Duarte:SPM:2008}, compressive spectral imaging system \cite{Gehm:OE:2007}, high-speed video camera \cite{Hitomi:ICCV:2011}, and fast Magnetic Resonance Imaging (MRI) system \cite{Lustig:SPM:2008} have been developed. Perfect reconstruction from the linear measurements is guaranteed if the following two conditions are met, i.e., the sensing matrices have the restricted isometry property (RIP) and the images have sparse representation with respect to a dictionary. These two conditions can be easily satisfied as random Gaussian matrices have the RIP with high probabilities \cite{Candes:IT:2006,Donoho:IT:2006} and natural images have sparse representation under many off-the-shelf or learned dictionaries (e.g., wavelets, learned dictionaries \cite{KSVD2006}). However, in practice the promise of CS is often offset by challenges relating to two conditions. First, it is difficult or even impossible to implement the random sensing matrix for a large image. Second, the sparsity-based CS reconstruction algorithms are very slow to converge for obtaining good estimates of the original images.

For a given image $\tx\in\mathbb{R}^{N}$, the CS measurement of $\tx$ can be expressed as $\ty=\tPhi$, where $\tPhi\in\mathbb{R}^{M\times N}$ is the sensing matrix and $M=rN$ ($0<r\ll1$). In practice, as the dimension of natural images are often very high, the memory requirement for storing the matrix $\tPhi$ is huge. For example, for a moderate image of size $1,000\times 1,000$ and $r=0.1$, the dimension of $\tPhi$ is $100,000\times 1,000,000$. The computational complexity of the CS reconstruction with such large sensing matrix is also prohibitively high. To avoid these difficulties, block-based CS (BCS) methods have been proposed \cite{BCS:FTSP12,BCS:ICIP09}, where an image is divided into many non-overlapping blocks and each block is sensed individually. As such, the sensing matrices $\tPhi$ become much smaller and images can be efficiently measured. However, as the blocks are sensed and reconstructed individually, the BCS methods lead to serious blocking artifacts. To reduce the blocking artifacts, post-processing is often required to improve the visual quality.

According to the CS theory \cite{Candes:IT:2006,Donoho:IT:2006}, the original images can be well-reconstructed by exploiting the sparsity prior of natural images. However, the sparsity-based CS methods recover the original images by solving an optimization algorithm, which is very slow to converge. Thus, the optimization-based CS methods cannot be used for real-time applications. Recently, inspired by the great successes of deep neural network (DNN) for computer vision tasks \cite{AlexNet:NIPS12,RCNN:CVPR14,FCN:CVPR15}, DNN-based CS reconstruction methods have also been proposed \cite{SDA:2015,ReconNet:2016,Elad:2017}. With DNN, both the linear sensing and nonlinear reconstruction can be performed by a single neural network. Through end-to-end training, both the sensing matrix and the reconstruction method can be jointly optimized, and the speed is usually hundreds-times faster than optimization-based methods. However, to the best of our knowledge, current DNN based methods are all block-based, i.e., image blocks are sensed and reconstructed individually. Blocking artifacts can also be observed in the reconstructed images by the DNN-based methods \cite{SDA:2015,ReconNet:2016}.

In this paper, we propose a novel convolutional compressive sensing (ConvCS) framework based on deep convolutional neural network (DCNN). In the proposed ConvCS network, the first layer senses an input image by convolving the \textit{whole image} with a set of random filters, followed by subsampling. The advantage of the proposed ConvCS is that the whole image can be efficiently sensed with a set of small filters that are easy to store, and effectively reconstructed without introducing blocking artifacts. The remaining layers of the proposed ConvCS network perform the nonlinear reconstruction of the \textit{whole image} from the measurements. To design the reconstruction network, the domain knowledge of the sparsity-based CS reconstruction is incorporated, leading to a novel CNN for image reconstruction. By end-to-end training, both the convolutional sensing filters and the reconstruction CNN are jointly optimized. Experimental results show that the proposed method substantially outperforms current state-of-the-art CS methods in terms of PSNR and visual quality.

\section{Related Works}

\subsection{Background of CS}
Instead of sampling the entire signal, the CS theory samples a signal $\tx\in\mathbb{R}^N$ by taking only $M$ linear measurements, i.e., $\ty = \tPhi\tx+\te$, $\ty\in\mathbb{R}^M$, where $\tPhi\in\mathbb{R}^{M\times N}$, $M<N$ is the sensing matrix and $\te\in\mathbb{R}^M$ is the measurement noise. Since $M<N$, recovering $\tx$ from $\ty$ is generally an ill-posed inverse problem. However, the CS theory guarantees perfect reconstruction of $\tx$ if $\tx$ is sparse in some sparsifying spaces and $\tPhi$ holds the RIP. It has been proven that the Gaussian random matrices have the RIP with very high probabilities. Standard CS methods recover $\tx$ by solving a minimization problem,
\begin{equation}
\valpha = \argmin_{\valpha} ||\ty-\tPhi\tD\valpha||_2^2 + \lambda||\valpha||_1,
\end{equation}
where $\tD$ is the dictionary and $\valpha$ are the sparse codes of $\tx$. After estimating the sparse codes $\valpha$, $\tx$ can be reconstructed as $\tx=\tD\valpha$. It has been proven in \cite{Candes:IT:2006,Donoho:IT:2006} that $\tx$ can be faithfully recovered from $M=O(K\log(N/K))$ measurements, where $K$ denotes the number of nonzero coefficients of $\valpha$. The $\ell_1$-minimization problem can be solved by many optimization algorithms \cite{ISTA:2004,Zibulevsky:2010}. However, as mentioned above these methods are very slow to converge.

\subsection{Block-based image CS}

When applying CS to images, the image can be sensed by representing it into a vector and measured as $\ty=\tPhi\tx$. However, for images of high dimensionality, the sensing matrix $\tPhi$ becomes very large and it is impossible to store it and compute with such large matrix $\tPhi$. To avoid such difficulties, block-based CS (BCS) methods have been proposed \cite{BCS:FTSP12,BCS:ICIP09}. In these methods, the input image is divided into many non-overlapped blocks and each block is sensed independently using a much smaller matrix $\tPhi$. The full image is recovered by placing back the reconstructed blocks, followed by full-image smoothing. To reduce the blocking artifacts, full-image iterative shrinkage algorithm was proposed \cite{BCS:ICIP09}, and improvements can be achieved by using more advanced transforms, such as contourlets and dual-tree discrete wavelet transforms \cite{BCS:FTSP12}. In addition to the sparsity prior, model-based CS recovery algorithms have also been developed to exploit the high-order dependencies between the wavelet coefficients \cite{baraniuk:IT:2010model}, leading to better performance. The effective nonlocal self-similarity prior has also been integrated into the objective function through nonlocal low-rank regularization \cite{Dong:TIP14}.

\subsection{Structured CS for images}
To overcome the drawback of the BCS, structured CS operators \cite{SIAM:2009,Yin:2010,LI:TSP13} have been proposed. The convolutional CS methods, which performs the sensing by first convoluting a signal with a random filter and then subsampling, have been proposed in \cite{SIAM:2009,LI:TSP13}. These CS methods are easy to implement and have many potential applications, such as Fourier optics \cite{SIAM:2009}, Radar imaging \cite{Yin:2010} and coded aperture imaging \cite{SPIE:2009}. In addition to the random filter, deterministic filter that is more convenient to implement has also been proposed in \cite{LI:TSP13}. Though the convolution-based CS are more easy to implement, the use of \textit{only one} random filter makes the reconstruction problem more difficult \cite{SIAM:2009}. Also, all these convolution-based CS use iterative algorithms to reconstruct the original images, which are very slow to converge.

\subsection{Deep learning based CS for images}
Recently, inspired by the successes of the deep neural networks, non-iterative CS reconstruction methods have been proposed \cite{SDA:2015,ReconNet:2016,Elad:2017}. In \cite{SDA:2015}, a stacked auto-encoder denoising network was developed for image CS. Similarly, a full-connected neural network has been proposed in \cite{Elad:2017}. Convolutional neural network has also been proposed for this task, where a BM3D denoising \cite{Dabov:TIP:2007} stage is adopted to further improve the reconstruction performance \cite{ReconNet:2016}. Through end-to-end training, both the sensing matrices and the reconstruction network can be jointly optimized for better performance. However, all those methods perform CS and reconstruction on image blocks, leading to limited reconstruction performance.

In this paper, we propose a new convolutional compressive sensing (denoted as ConvCS) framework using deep convolutional neural network (DCNN), where the first layer implements the sensing by convolving the input image with \textit{a set} of random filters followed by subsampling. The remaining layers reconstruct the input image from the linear measurements. Furthermore, inspired from the sparsity-based reconstruction model, a novel CNN containing two branches is proposed for CS reconstruction. By performing sensing and reconstruction on the \textit{whole} image, the proposed method significantly outperforms the previous CS methods. Different from previous convolution-based CS methods \cite{SIAM:2009,LI:TSP13}, both the sensing filters and the reconstruction algorithm can be jointly optimized. Experimental results show that the proposed method outperforms existing state-of-the-art CS methods by a large margin.

\section{Proposed Convolutional Compressive Sensing using Deep Learning}

\subsection{Proposed convolutional compressive sensing}

Unlike existing BCS and convolutional CS, we propose to sense an image by convolving it with a set of random filters, followed by spatial subsampling of the convolved images. Specifically, for a given image $\tx\in\mathbb{R}^N$, we convolve it with a set of random filters $\vphi_i$ of size $L\times L$, $i=1,2,\cdots,m$, to generate the CS measurements. Mathematically, the sensing matrix can be expressed as
\begin{equation}
\tPhi = [(\tS\tPhi_1)^{\top}, (\tS\tPhi_2)^{\top}, \cdots, (\tS\tPhi_m)^{\top}]^{\top}, \label{ConvCS}
\end{equation}
where $\tPhi_i\in\mathbb{R}^{N\times N}$ is set to be a sparse matrix such that $\tPhi_i\tx$ is equivalent to convolving $\tx$ with filter $\vphi_i$, and $\tS\in\mathbb{R}^{M_0\times N}$ is a subsampling matrix. Then, the proposed convolutional CS (ConvCS) for $\tx$ can be formulated as $\ty=\tPhi\tx$, $\ty\in\mathbb{R}^{M}$, where $M=mM_0$.

As convolution can also be implemented with matrix-vector multiplication, the proposed ConvCS is equivalent to the CS process that first extracts blocks of size $L\times L$ with sliding step $s$ and then measures the blocks with the sensing matrix, whose $i$-th rows are composed with the vectorized filter coefficients of $\vphi_i$. Thus, the proposed ConvCS matrix $\tPhi$ of Eq. (\ref{ConvCS}) still holds the advantages of the random Gaussian matrix for CS (i.e., the RIP). However, the proposed ConvCS is clearly distinct from the previous BCS methods in two aspects. First, the proposed ConvCS for images of large dimensions is much easier to implement. Second, the convolutional nature of the proposed ConvCS makes the joint optimization of the sensing filters $\vphi_i$ and reconstruction of the \textit{whole} image much more effective, without introducing any blocking artifacts.

\begin{figure*}[!tbh]
\centering
    {\includegraphics[width=0.7\linewidth,height=0.25\linewidth]{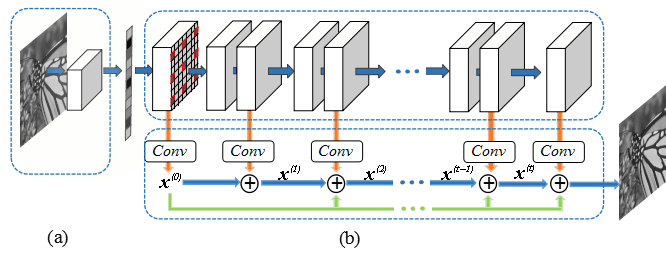}}
\caption{The proposed convolutional compressive sensing network (ConvCSNet). (a) Convolutional compressive sensing layer; (b) Two-branch convolutional neural network for CS reconstruction.}
\label{fig:ConvCS}
\end{figure*}

The proposed ConvCS can be easily implemented using a convolutional neural network, as shown in Fig. \ref{fig:ConvCS} (a). The first layer convolves the input image $\tx$ with a set of $m$ random filters of size $L\times L$ with stride $s$. The measurements can be obtained by representing the obtained feature maps into a $M$-dimensional vector $\ty$. Note that nonlinearity is not involved in the CS process. More convolutional layers with nonlinear activity function could be added into the ConvCS process, which may lead to better performance. However, for simplicity, here we only use one linear layer to obtain the measurements. As shown in Sec. \ref{Exp_sec}, the simple ConvCS can already lead to excellent CS recovery performances.

Essentially, the ConvCS encodes the visual information of images and plays a similar role as autoencoders. In the context of CS, the stacked denoising autoencoders (SDA) \cite{SDA:2015} have been proposed to encode the images. However, the SDA method performs CS encoding at image block levels, resulting in serious blocking artifacts.


\subsection{Sparse model inspired DCNN for CS reconstruction}
\label{sub_recon}

After obtaining the CS measurements, we aim to recover the original image from the measurements. In recent years, DCNN has shown very promising performances for many low-level image processing tasks, e.g., image super-resolution \cite{SRCNN:2014,VDSR:2016}. However, the problem of reconstructing images from the ConvCS measurements is different from previous image SR problem, and existing reconstruction network can not be applied directly for this task. To facilitate the design of the CS reconstruction network, we propose to incorporate the domain knowledge of the sparsity-based CS reconstruction. Specifically, we first propose the following analysis sparse representation model for CS reconstruction,
\begin{equation}
\min_{\tx,\valpha_k} ||\ty - \tPhi\tx||_2^2 + \eta\sum_k^K \{||\tw_k*\tx - \valpha_k||_F^2 + J(\valpha_k) \}, \label{CS_obj}
\end{equation}
where $\tw_k\in\mathbb{R}^{n\times n}$ is the analysis filter, $*$ denotes the 2D convolution, and $J(\cdot)$ denotes a regularization term imposed on sparse codes $\valpha_k$. Classic sparsity enforcing regularizers, e.g., $||\cdot||_p$ ($0\leq p\leq 1$), as well as the non-negative indicator function $\mathcal{I}(\cdot)$ (as inspired by the ReLU function) can be adopted. Solving Eq. (\ref{CS_obj}) amounts to alternatively solving two subproblems, i.e.,
\begin{equation}
\begin{split}
&\tx =\argmin_{\tx} ||\ty-\tPhi\tx||_2^2 + \eta\sum_k^K \{||\tw_k*\tx-\valpha_k||_F^2 \}, \\
&\valpha_k = \argmin_{\valpha_k} ||\tw_k*\tx-\valpha_k||_F^2 + J(\valpha_k),
\end{split}
\end{equation}
where $k=1,2,\cdots,K$. Both sub-problems can be easily optimized. With a fixed estimate $\tx^{(t)}$ obtained at $t$-th iteration, $\valpha_k$ can be solved in closed-form for several commonly used sparsity regularizer, as
\begin{equation}
\valpha_k^{(t+1)} = \mathcal{S}_{\tau} (\tw_k*\tx^{(t)}), \text{or} ~ [\tw_k*\tx^{(t)}]_+ \label{alpha_update}
\end{equation}
for $\ell_1$-norm sparse regularizer and non-negative regularizer, respectively, where $\mathcal{S}_{\tau}(\cdot)$ denotes the soft-thresholding with threshold $\tau$. The $x$-subproblem is a quadratic optimization problem and can be solved in closed-form. However, to avoid large matrix inversion, we prefer to solve it via a gradient descent method. With a fixed estimate of $\valpha_k$, we iteratively update $\tx$ as
\begin{equation}
\begin{split}
\tx^{(t+1)}  =  &\tx^{(t)} - \delta(\tPhi^{\top}(\tPhi\tx^{(t)}-\ty) + \\
&\eta\sum_k^K (\tW_k^{\top}(\tW_k\tx^{(t)}-\valpha_k^{(t+1)}))), \label{x-update}
\end{split}
\end{equation}
where $\tW_k$ is the sparse matrix such that $\tW_k\tx$ is equivalent to convolving $\tx$ with $\tw_k$, and $\delta$ is the predefined constant. By alternatively updating $\tx$ and $\valpha$, the iterative process will converge. However, the convergence speed is very slow.

In this paper, we propose to convert the alternative update of $\tx$ and $\valpha$ into a deep network. For simplicity, we rewrite Eq. (\ref{x-update}) as
\begin{equation}
\tx^{(t+1)} = \tx^{(t)}-\delta(\tLambda\tx^{(t)}-\tPhi^{\top}\ty + \eta \sum_k^K \tW_k^{\top}(\tW_k\tx^{(t)}-\valpha_k^{(t+1)})), \label{x-update2}
\end{equation}
where $\tLambda\tx=\tPhi^{\top}\tPhi\tx$ denotes the process that first senses $\tx$ with $\tPhi$ and then reconstructs $\tx$ from the CS measurements by back-projection. For simplicity, we let $\tLambda=\tI$ and $\tW_k^{\top}\tW_k=\tI$. Then, Eq. (\ref{x-update2}) can be approximated as
\begin{equation}
\tx^{(t+1)} \approx \rho\tx^{(t)} + \delta\tx^{(0)} + \gamma \tx^{(t+1/2)},\label{x-update3}
\end{equation}
where $\rho=(1-\delta(1+\eta))$, $\gamma=\delta\eta$, $\tx^{(0)}$ denotes the initial reconstruction of $\tx$ from $\ty$, and $\tx^{(t+1/2)}=\sum_k^K\tW_k^{\top}\valpha_k^{(t+1)}$ denotes the reconstructed $\tx$ from $\valpha_k^{(t+1)}$.

Inspired from the alternative update of $\valpha_k$ and $\tx$, we propose a novel DCNN for CS reconstruction. As shown in Fig. \ref{fig:ConvCS} (b), the proposed reconstruction network contains two branches. The first branch implements a conventional CNN for generating the feature maps $\valpha_k$. The first branch taking the measurements vector $\ty$ as input back projects $\ty$ into the feature maps, where all entries are zero except the sampled set of pixels (as marked red shown in Fig. \ref{fig:ConvCS} (b)). The first layer uses kernels of size $L\times L$ and generates $m$ feature maps, while the remaining $14$ layers use kernels of size $3\times 3$ and generate $96$ feature maps. The ReLU function is applied following convolution. Compared to the directly implementation of Eq. (\ref{alpha_update}), the deep CNN is more powerful in learning the representation of the original image $\tx$. The first CNN branch can also be regarded as a nonlinear mapping function used to accurately predict the sparse codes $\valpha_k$. The second branch recursively reconstructs the image based on the feature maps $\valpha$ from the first branch and the previously reconstructed images $\tx^{(t)}$ and $\tx^{(0)}$, which mimics the computations of Eqs. (\ref{x-update2}) and (\ref{x-update3}). In each layer of the second branch, the feature map $\valpha^{(t+1)}$ from the CNN branch is fed into a convolutional layer to produce an image, which is further added with previously reconstructed $\tx^{(t)}$ and $\tx^{(0)}$ for an updated estimate $\tx^{(t+1)}$. The kernel size used in the reconstruction branch is also $3\times 3$.


\section{Experimental results}
\label{Exp_sec}
\subsection{Training details}

To achieve better performance, the sensing layers and the remaining reconstruction layers are jointly trained, and thus the sensing filters $\vphi_i$ and the reconstruction network can be jointly optimized. Let ConvCSNet denote the proposed network performing convolutional CS and reconstruction. To verify the effectiveness of the proposed two-branch reconstruction network, we also implement a variant of the ConvCSNet (denoted as ConvCSNet-baseline), which uses only the first branch of the reconstruction network shown in Fig. \ref{ConvCS} (b) for CS reconstruction. To train the proposed networks, we collected natural images from the ImageNet dataset \cite{ImageNet:IJCV14} and extracted the central $160\times 160$ part of each image. The extracted patches are converted into grayscale and augmented via horizontal and vertical flips and $90$ rotations. Finally, we obtained a training set containing $4000,000$ image patches.  We empirically set the parameters of the convolutional sensing layer as: $L=17, m=8, s=8$ for rate $0.05$; $L=17, m=6, s=8$ for rate $0.1$, $L=11, m=5, s=5$ for rate $0.2$, and $L=11, m=8, s=5$ for rate $0.3$, where $L, m$ and $s$ denotes the filter size, number of filters and the convolutional stride, respectively. The parameters of the reconstruction layers are the same for different measurement rates, except the first layer of the reconstruction part of the ConvCSNet. In that layer, we use corresponding filter size of $L\times L$ used in the sensing layer. 

The proposed network is trained using the $\ell_2$ loss function, as $L(\vTheta)=\frac{1}{T}\sum_i^T||f(\tx_i,\vTheta)-\tx_i||_2^2$, where $\vTheta$ denotes all the network parameters (including the sensing filters) and $T$ is the total number of training patches. The proposed network is trained with ADAM optimizer \cite{ADAM:ICLR14} by setting $\beta_1=0.9$, $\beta_2=0.999$ and $\epsilon=10^{-8}$. The minibatch size is set to $64$. We initialize the learning rate as $10^{-4}$ and halved at every $2\times 10^5$ minibatch updates. We implemented the proposed network with TensorFlow framework and training them using 4 NVIDIA 1080Ti GPUs. It takes one day to train ConvCSNet. Currently, we trained one model for each sensing rates. In future, a general reconstruction network may be trained for different measurement rates.

\subsection{Comparison with state-of-the-art methods }

\begin{figure}
\renewcommand{\arraystretch}{0.4}
\centering
   \includegraphics[width=0.13\linewidth]{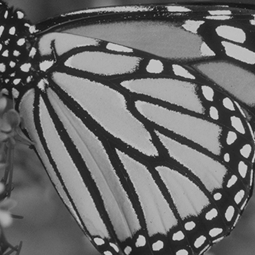}
   \includegraphics[width=0.13\linewidth]{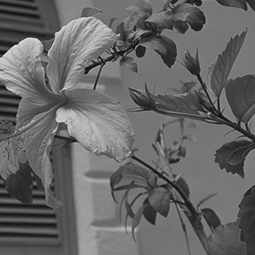}
   \includegraphics[width=0.13\linewidth]{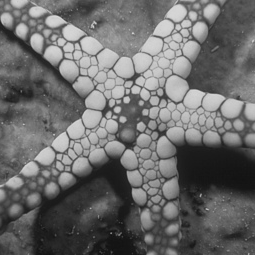}
   \includegraphics[width=0.13\linewidth]{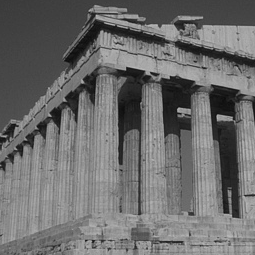}
   \includegraphics[width=0.13\linewidth]{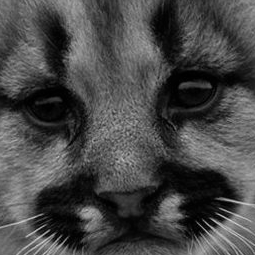}
   \includegraphics[width=0.13\linewidth]{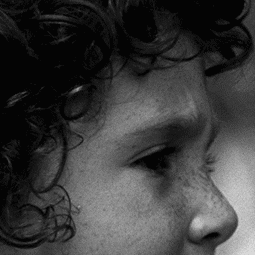}
   \includegraphics[width=0.13\linewidth]{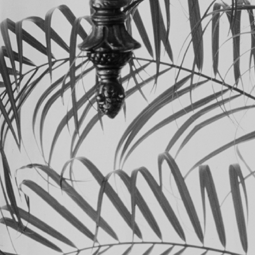}
   \caption{The test images. From left to right: \textit{Butterfly, Flowers, Starfish, Parthenon, raccoon, Girl}, and \textit{Leaves}.}
\label{fig:TestImages}
\end{figure}

\begin{table*}[]
\centering
\caption{PSNR results on the set of images of Fig. \ref{fig:TestImages} by the test methods at different measurement rates.}
\label{PSNR_Tab1}
\begin{tabular}{|c|c|c|c|c|c|c|c|c|c|}
\hline
\multirow{2}{*}{Images} &                                                                      & \multicolumn{4}{c|}{Noiseless}                                                                                                                                                                                                                                                                                                    & \multicolumn{4}{c|}{Noisy}                                                                                                                                                                                                                                                                                                    \\ \cline{2-10}
                        & Ratio                                                                & 0.05                                                                          & 0.1                                                                           & 0.2                                                                           & 0.3                                                                           & 0.05                                                                          & 0.1                                                                           & 0.2                                                                           & 0.3                                                                           \\ \hline
butterfly               & \begin{tabular}[c]{@{}c@{}}TV\cite{LI:COA}\\ FCN-CS\cite{Elad:2017}\\ D-AMP\cite{AMP:TIF16}\\ \textbf{ConvCSNet-baseline}\\ \textbf{ConvCSNet}\end{tabular} & \begin{tabular}[c]{@{}c@{}}19.01\\ 22.93\\ 22.70\\ 24.17\\ \textbf{25.63}\end{tabular} & \begin{tabular}[c]{@{}c@{}}22.73\\ 25.78\\ 26.55\\ 27.50\\ \textbf{28.48}\end{tabular} & \begin{tabular}[c]{@{}c@{}}28.03\\ 29.07\\ 31.38\\ 31.44\\ \textbf{32.91}\end{tabular} & \begin{tabular}[c]{@{}c@{}}31.99\\ 30.87\\ 34.32\\ 34.72\\ \textbf{35.36}\end{tabular} & \begin{tabular}[c]{@{}c@{}}18.56\\ 22.82\\ 20.52\\ 23.99\\ \textbf{25.35}\end{tabular} & \begin{tabular}[c]{@{}c@{}}21.88\\ 24.83\\ 23.40\\ 27.11\\ \textbf{28.46}\end{tabular} & \begin{tabular}[c]{@{}c@{}}25.11\\ 27.01\\ 27.75\\ 29.85\\ \textbf{30.70}\end{tabular} & \begin{tabular}[c]{@{}c@{}}26.28\\ 29.40\\ 28.93\\ 32.32\\ \textbf{33.49}\end{tabular} \\ \hline
parthenon               & \begin{tabular}[c]{@{}c@{}}TV\cite{LI:COA}\\ FCN-CS\cite{Elad:2017}\\ D-AMP\cite{AMP:TIF16}\\ \textbf{ConvCSNet-baseline}\\ \textbf{ConvCSNet}\end{tabular} & \begin{tabular}[c]{@{}c@{}}22.75\\ 25.28\\ 24.59\\ 25.34\\ \textbf{25.79}\end{tabular} & \begin{tabular}[c]{@{}c@{}}24.90\\ 27.18\\ \textbf{27.43}\\ 26.80\\ 27.39\end{tabular} & \begin{tabular}[c]{@{}c@{}}27.28\\ 29.93\\ 30.72\\ 30.01\\ \textbf{30.94}\end{tabular} & \begin{tabular}[c]{@{}c@{}}29.18\\ \textbf{31.96}\\ 33.11\\ 33.35\\ 33.89\end{tabular} & \begin{tabular}[c]{@{}c@{}}22.16\\ 25.21\\ 23.78\\ 24.99\\ \textbf{25.65}\end{tabular} & \begin{tabular}[c]{@{}c@{}}23.76\\ 26.59\\ 25.50\\ 26.46\\ \textbf{27.31}\end{tabular} & \begin{tabular}[c]{@{}c@{}}25.21\\ 28.58\\ 27.62\\ 28.70\\ \textbf{29.53}\end{tabular} & \begin{tabular}[c]{@{}c@{}}25.95\\ 30.49\\ 28.90\\ 31.23\\ \textbf{32.44}\end{tabular} \\ \hline
starfish                & \begin{tabular}[c]{@{}c@{}}TV\cite{LI:COA}\\ FCN-CS\cite{Elad:2017}\\ D-AMP\cite{AMP:TIF16}\\ \textbf{ConvCSNet-baseline}\\ \textbf{ConvCSNet}\end{tabular} & \begin{tabular}[c]{@{}c@{}}22.07\\ 24.99\\ 23.87\\ 25.04\\ \textbf{25.79}\end{tabular} & \begin{tabular}[c]{@{}c@{}}24.31\\ 27.35\\ 27.44\\ 27.29\\ \textbf{28.79}\end{tabular} & \begin{tabular}[c]{@{}c@{}}27.58\\ 30.91\\ 31.73\\ 32.61\\ \textbf{34.10}\end{tabular} & \begin{tabular}[c]{@{}c@{}}30.24\\ 33.25\\ 34.46\\ 36.17\\ \textbf{37.51}\end{tabular} & \begin{tabular}[c]{@{}c@{}}21.65\\ 24.94\\ 22.71\\ 24.62\\ \textbf{25.67}\end{tabular} & \begin{tabular}[c]{@{}c@{}}23.50\\ 27.16\\ 25.20\\ 26.94\\ \textbf{28.72}\end{tabular} & \begin{tabular}[c]{@{}c@{}}25.37\\ 29.76\\ 27.82\\ 30.27\\ \textbf{31.77}\end{tabular} & \begin{tabular}[c]{@{}c@{}}26.57\\ 32.11\\ 29.25\\ 33.19\\ \textbf{35.00}\end{tabular} \\ \hline
flower                  & \begin{tabular}[c]{@{}c@{}}TV\cite{LI:COA}\\ FCN-CS\cite{Elad:2017}\\ D-AMP\cite{AMP:TIF16}\\ \textbf{ConvCSNet-baseline}\\ \textbf{ConvCSNet}\end{tabular} & \begin{tabular}[c]{@{}c@{}}23.48\\ 26.00\\ 24.96\\ 26.10\\ \textbf{26.80}\end{tabular} & \begin{tabular}[c]{@{}c@{}}25.81\\ 28.08\\ 28.34\\ 27.88\\ \textbf{29.14}\end{tabular} & \begin{tabular}[c]{@{}c@{}}28.85\\ 30.91\\ 32.49\\ 32.14\\ \textbf{33.38}\end{tabular} & \begin{tabular}[c]{@{}c@{}}31.33\\ 32.83\\ 36.18\\ 35.67\\ \textbf{36.11}\end{tabular} & \begin{tabular}[c]{@{}c@{}}22.9\\ 25.92\\ 24.07\\ 25.72\\ \textbf{26.60}\end{tabular}  & \begin{tabular}[c]{@{}c@{}}24.39\\ 27.81\\ 26.13\\ 27.45\\ \textbf{29.07}\end{tabular} & \begin{tabular}[c]{@{}c@{}}26.03\\ 29.90\\ 28.20\\ 30.40\\ \textbf{31.45}\end{tabular} & \begin{tabular}[c]{@{}c@{}}26.88\\ 31.83\\ 29.64\\ 32.82\\ \textbf{34.49}\end{tabular} \\ \hline
girl                    & \begin{tabular}[c]{@{}c@{}}TV\cite{LI:COA}\\ FCN-CS\cite{Elad:2017}\\ D-AMP\cite{AMP:TIF16}\\ \textbf{ConvCSNet-baseline}\\ \textbf{ConvCSNet}\end{tabular} & \begin{tabular}[c]{@{}c@{}}28.78\\ 31.29\\ 30.35\\ 31.55\\ \textbf{31.97}\end{tabular} & \begin{tabular}[c]{@{}c@{}}30.41\\ 32.87\\ 31.87\\ 33.10\\ \textbf{33.43}\end{tabular} & \begin{tabular}[c]{@{}c@{}}32.20\\ 34.58\\ 33.43\\ 34.69\\ \textbf{35.22}\end{tabular} & \begin{tabular}[c]{@{}c@{}}33.47\\ 35.76\\ 34.42\\ 34.5\\ \textbf{36.55}\end{tabular}  & \begin{tabular}[c]{@{}c@{}}27.00\\ 31.24\\ 28.18\\ 31.05\\ \textbf{31.57}\end{tabular} & \begin{tabular}[c]{@{}c@{}}27.78\\ 32.68\\ 29.69\\ 32.22\\ \textbf{33.17}\end{tabular} & \begin{tabular}[c]{@{}c@{}}28.31\\ \textbf{33.95}\\ 30.87\\ 32.72\\ 33.34\end{tabular} & \begin{tabular}[c]{@{}c@{}}28.46\\ 35.02\\ 31.59\\ 33.64\\ \textbf{34.14}\end{tabular} \\ \hline
leaves                  & \begin{tabular}[c]{@{}c@{}}TV\cite{LI:COA}\\ FCN-CS\cite{Elad:2017}\\ D-AMP\cite{AMP:TIF16}\\ \textbf{ConvCSNet-baseline}\\ \textbf{ConvCSNet}\end{tabular} & \begin{tabular}[c]{@{}c@{}}17.64\\ 22.13\\ 18.69\\ 22.09\\ \textbf{23.51}\end{tabular} & \begin{tabular}[c]{@{}c@{}}20.10\\ 24.67\\ 26.80\\ 24.76\\ \textbf{27.03}\end{tabular} & \begin{tabular}[c]{@{}c@{}}24.50\\ 28.82\\ 33.00\\ 30.93\\ \textbf{33.32}\end{tabular} & \begin{tabular}[c]{@{}c@{}}28.08\\ 31.3\\ 36.77\\ 34.73\\ \textbf{36.98}\end{tabular}  & \begin{tabular}[c]{@{}c@{}}17.27\\ 22.06\\ 20.20\\ 21.82\\ \textbf{23.38}\end{tabular} & \begin{tabular}[c]{@{}c@{}}19.81\\ 24.31\\ 24.42\\ 24.44\\ \textbf{27.05}\end{tabular} & \begin{tabular}[c]{@{}c@{}}23.02\\ 27.27\\ 27.77\\ 29.18\\ \textbf{30.96}\end{tabular} & \begin{tabular}[c]{@{}c@{}}24.94\\ 30.12\\ 29.2\\ 33.15\\ \textbf{35.20}\end{tabular}  \\ \hline
raccoon                 & \begin{tabular}[c]{@{}c@{}}TV\cite{LI:COA}\\ FCN-CS\cite{Elad:2017}\\ D-AMP\cite{AMP:TIF16}\\ \textbf{ConvCSNet-baseline}\\ \textbf{ConvCSNet}\end{tabular} & \begin{tabular}[c]{@{}c@{}}24.08\\ 26.17\\ 25.16\\ 25.68\\ \textbf{26.27}\end{tabular} & \begin{tabular}[c]{@{}c@{}}25.66\\ 27.63\\ 27.12\\ 27.12\\ \textbf{27.95}\end{tabular} & \begin{tabular}[c]{@{}c@{}}27.53\\ 29.82\\ 29.44\\ 30.73\\ \textbf{31.20}\end{tabular} & \begin{tabular}[c]{@{}c@{}}29.12\\ 31.66\\ 31.77\\ 33.38\\ \textbf{34.17}\end{tabular} & \begin{tabular}[c]{@{}c@{}}23.60\\ \textbf{26.16}\\ 24.34\\ 25.44\\ 26.14\end{tabular} & \begin{tabular}[c]{@{}c@{}}24.60\\ 27.53\\ 25.87\\ 26.73\\\textbf{27.87}\end{tabular} & \begin{tabular}[c]{@{}c@{}}25.83\\ 29.43\\ 27.48\\ 29.45\\ \textbf{29.71}\end{tabular} & \begin{tabular}[c]{@{}c@{}}26.47\\ 31.25\\ 28.47\\ 31.29\\ \textbf{32.58}\end{tabular} \\ \hline
Average                    & \begin{tabular}[c]{@{}c@{}}TV\cite{LI:COA}\\ FCN-CS\cite{Elad:2017}\\ D-AMP\cite{AMP:TIF16}\\ \textbf{ConvCSNet-baseline}\\ \textbf{ConvCSNet}\end{tabular} & \begin{tabular}[c]{@{}c@{}}22.54\\ 25.54\\ 24.33\\ 25.71\\ \textbf{26.54}\end{tabular} & \begin{tabular}[c]{@{}c@{}}24.85\\ 27.65\\ 27.94\\ 27.78\\ \textbf{28.89}\end{tabular} & \begin{tabular}[c]{@{}c@{}}28.00\\ 30.58\\ 31.74\\ 31.79\\ \textbf{33.01}\end{tabular} & \begin{tabular}[c]{@{}c@{}}30.49\\ 32.52\\ 34.43\\ 34.65\\ \textbf{35.80}\end{tabular} & \begin{tabular}[c]{@{}c@{}}21.88\\ 25.48\\ 23.40\\ 25.38\\ \textbf{26.34}\end{tabular} & \begin{tabular}[c]{@{}c@{}}23.68\\ 27.27\\ 25.96\\ 27.34\\ \textbf{28.81}\end{tabular} & \begin{tabular}[c]{@{}c@{}}25.55\\ 29.41\\ 28.22\\ 30.08\\ \textbf{31.07}\end{tabular} & \begin{tabular}[c]{@{}c@{}}26.51\\ 31.46\\ 29.43\\ 32.52\\ \textbf{33.91}\end{tabular} \\ \hline
\end{tabular}
\end{table*}

We compare the proposed ConvCSNet method with two iterative CS methods, including the total variation method (denoted as TV \cite{LI:COA}), the denoising-based approximate message passing (D-AMP) method \cite{AMP:TIF16}, and one recently developed fully-connected network based CS method (denoted as FCN-CS) \cite{Elad:2017}. Note that both TV \cite{LI:COA} and D-AMP methods \cite{AMP:TIF16} conduct compressive sensing on the \textit{whole image}, using very large measurement matrices. For an input image of size $255\times 255$ and measurement rate $0.2$, the measurement matrix is of size $13005 \times 65032$, requiring about $7 GB$ memory for storing it in Matlab platform. The computational complexity of the iterative CS reconstruction using such large sensing matrices also become very slow. However, the advantage of using such large measurement matrices is that the reconstruction quality can be much improved. Also note that the D-AMP method \cite{AMP:TIF16} uses the well-known BM3D denoising method \cite{Dabov:TIP:2007} in its iterative reconstruction process. As the BM3D denoising method is very effective in suppressing noise and artifacts, the D-AMP method achieves the state-of-the-art CS performance. The deep learning based FCN-CS method \cite{Elad:2017} performing the sensing and reconstruction based on $16\times 16$ blocks. As the authors of FCN-CS method only provided the test code in their website, we re-implemented the training algorithm of FCN-CS method and trained the network with our training dataset. As we use larger training dataset, the performance of FCN-CS method is much improved, compared with those obtained using the model provided by the authors of \cite{Elad:2017}.

We have also tried to compare our method with the ReconNet of \cite{ReconNet:2016}. However, the results obtained using the code downloaded from their website are worse than those reported in their paper. We think that this may be caused by different parameters settings, and tuning the parameters of the method for better results is out of the scope of this paper. Hence, we didn't include the ReconNet \cite{ReconNet:2016} into our comparison study. All the codes of the competing methods are downloaded from authors' websites. We generate the measurements at four measurement rates $0.05, 0.1, 0.2$, and $0.3$. To verify the robustness of the reconstruction algorithms to the measurement noise, we also conducted CS reconstruction using noisy measurements. To this end, Gaussian noise of standard deviation of $10$ is added to the measurements. A set of natural images of size $255\times 255$ is used as test images, as shown in Fig. \ref{fig:TestImages}. The well-known Berkeley segmentation dataset containing 100 natural images (denoted as BSD100) is also used to verify the performances of the test methods. For the BSD100 dataset, we extract the central $320\times 320$ part of each image as test images. Note that all the test images are excluded from the training dataset.

\begin{table*}[]
\centering
\caption{Average PSNR results of different methods on test set of BSD100.}
\label{PSNR_Tab2}
\begin{tabular}{|c|c|c|c|c|c|c|c|c|}
\hline
      & \multicolumn{4}{c|}{Noiseless}    & \multicolumn{4}{c|}{Noisy}   \\ \hline
Ratio & 0.05  & 0.1   & 0.2   & 0.3   & 0.05  & 0.1   & 0.2   & 0.3  \\ \hline
TV\cite{LI:COA}    & 24.09 & 26.00    & 28.46 & 30.45    & 21.96 & 22.95 & 24.17    & 25.16  \\ \hline
FCN-CS\cite{Elad:2017}   & 25.92    & 27.54 & 29.85 & 31.57 & 25.89    & 27.29    & 29.03   & 30.74   \\ \hline
D-AMP\cite{AMP:TIF16}  & 25.48 & 27.66 & 30.65 &33.12    & 22.63 & 23.72 & 24.95 & 25.81   \\ \hline
\textbf{ConvCSNet-baseline}  & 26.25 & 28.12 & 30.36 & 33.08    & 25.80 & 27.66 & 28.96 & 31.15   \\ \hline
\textbf{ConvCSNet}  & \textbf{26.47} & \textbf{28.19}  & \textbf{31.03} & \textbf{33.49} & \textbf{26.22}  & \textbf{28.09} & \textbf{29.63} & \textbf{32.25} \\ \hline
\end{tabular}
\end{table*}

\begin{figure*}[!tbh]
\renewcommand{\arraystretch}{0.4}
\centering
\subfigure{
    \includegraphics[width=0.19\textwidth]{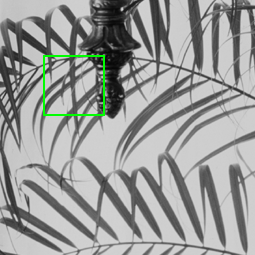}
    }\hspace{-0.8em}
\subfigure{
   \includegraphics[width=0.19\textwidth]{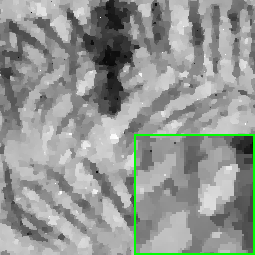}
    }\hspace{-0.8em}
\subfigure{
   \includegraphics[width=0.19\textwidth]{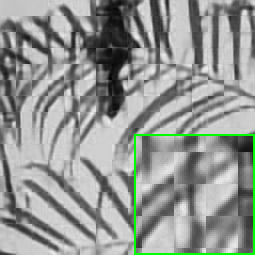}
    }\hspace{-0.8em}
\subfigure{
   \includegraphics[width=0.19\textwidth]{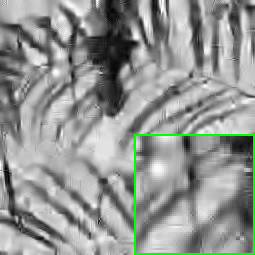}
    }\hspace{-0.8em}
\subfigure{
   \includegraphics[width=0.19\textwidth]{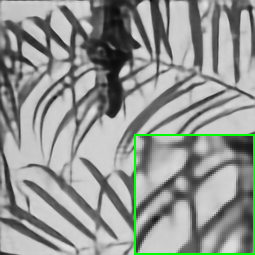}
    }
\\
\setcounter{subfigure}{0}
\subfigure[Original]{
    \includegraphics[width=0.19\textwidth]{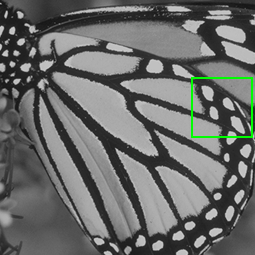}
    }\hspace{-0.8em}
\subfigure[TV\cite{LI:COA}]{
   \includegraphics[width=0.19\textwidth]{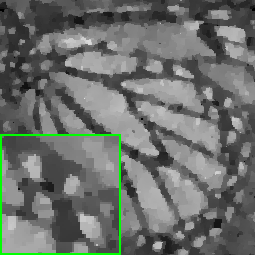}
    }\hspace{-0.8em}
\subfigure[FCN-CS\cite{Elad:2017}]{
   \includegraphics[width=0.19\textwidth]{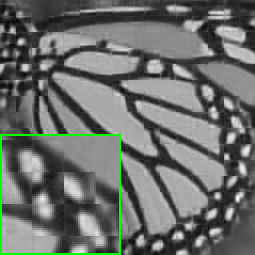}
    }\hspace{-0.8em}
\subfigure[D-AMP\cite{AMP:TIF16}]{
   \includegraphics[width=0.19\textwidth]{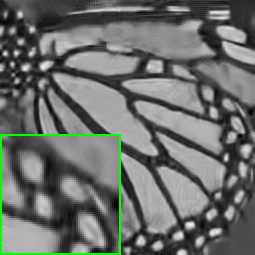}
    }\hspace{-0.8em}
\subfigure[\textbf{Proposed ConvCSNet} ]{
   \includegraphics[width=0.19\textwidth]{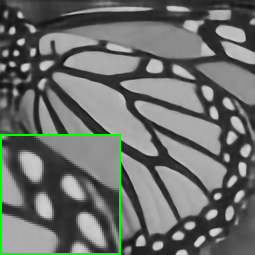}
    }
     \caption{Reconstructed results for \textit{Leaves} and \textit{Butterfly} images from noiseless CS measurements at sensing rate $0.05$. The PSNR results: (b) TV \cite{LI:COA} (17.64 dB, 19.01 dB); (c)FCN-CS\cite{Elad:2017} (22.13 dB, 22.93 dB); (d)D-AMP\cite{AMP:TIF16} (18.69 dB, 22.70 dB); (e) Proposed ConvCSNet (\textbf{23.51} dB, \textbf{25.63} dB) }
  \label{fig: butterfly_0.05_clean}
\end{figure*}

\begin{figure*}[!tbh]
\renewcommand{\arraystretch}{0.4}
\centering
\subfigure{
    \includegraphics[width=0.19\textwidth]{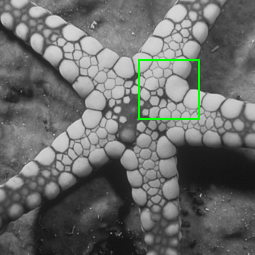}
    }\hspace{-0.8em}
\subfigure{
   \includegraphics[width=0.19\textwidth]{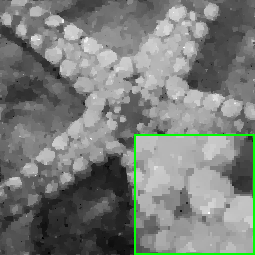}
    }\hspace{-0.8em}
\subfigure{
   \includegraphics[width=0.19\textwidth]{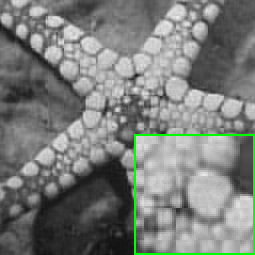}
    }\hspace{-0.8em}
\subfigure{
   \includegraphics[width=0.19\textwidth]{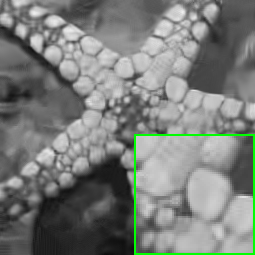}
    }\hspace{-0.8em}
\subfigure{
   \includegraphics[width=0.19\textwidth]{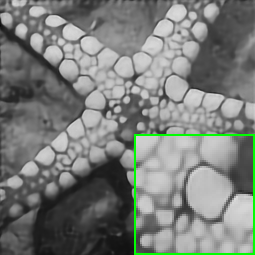}
    }\hspace{-0.8em}
     \caption{Reconstructed results for \textit{Starfish} images from noisy CS measurements at sensing rate $0.1$. The PSNR results: (b) TV \cite{LI:COA} (23.50 dB); (c)FCN-CS\cite{Elad:2017} (27.16 dB); (d)D-AMP\cite{AMP:TIF16} (25.20 dB); (e) Proposed ConvCSNet (\textbf{28.72} dB) }
  \label{fig: starfish_0.1_noisy}
\end{figure*}


\begin{table}[]
\centering
\setlength{\tabcolsep}{1.6mm}
\small
\caption{Running time (second) of different methods on a test image of size $255\times255$ with measurement rate 0.2.}
\label{Table_time}
\begin{tabular}{|c|c|c|c|c|c|}
\hline
Methods     &TV\cite{LI:COA}    & FCN-CS\cite{Elad:2017}  & D-AMP\cite{AMP:TIF16} & \textbf{ConvCSNet} \\ \hline
Time   &82.47 & 0.41 & 60.79  & 0.08 \\ \hline
\end{tabular}
\end{table}

Table \ref{PSNR_Tab1} reports the PSNR results of the test methods on the set of test images shown in Fig. \ref{fig:TestImages}. It can be seen that the proposed ConvCSNet outperforms the ConvCSNet-baseline method by a large margin, demonstrating the effectiveness of the proposed CS reconstruction network. When compared with other competing methods, the proposed ConvCSNet method also performs better than the other methods on all measurement rates. It significantly outperforms the block-based FCN-CS method on all measurement rates. The D-AMP method performs very well for high measurement rates (e.g., 0.2 and 0.3). The proposed ConvCSNet also performs much better than D-AMP method on all measurement rates. For noisy cases, the proposed ConvCSNet also significantly outperforms the other methods by large margins. Table \ref{PSNR_Tab2} shows the average PSNR results of the test methods on the BSD100 dataset. From Table \ref{PSNR_Tab2}, we can see that the proposed ConvCSNet also outperforms all the other competing methods, for both noiseless and noisy cases.

Figs. \ref{fig: butterfly_0.05_clean} and \ref{fig: starfish_0.1_noisy} show parts of the reconstructed images by the test methods. Clearly, the visual quality of the images reconstructed by the proposed ConvCSNet method is significantly better than other competing methods. The proposed method can generate visually pleasant images at measurement rate $0.05$, while other methods produce images with severe visual artifacts. For more visual comparisons, please refer to the supplementary material.

Regarding the computational complexity of the test methods, we report the running time of the test methods on a test image of size $255\times 255$ for the noiseless case, as shown in Table \ref{Table_time}. For TV \cite{LI:COA} and D-AMP \cite{AMP:TIF16} methods, the computer with Intel i7-6700 3.4GHz and $64$G memory is used to run these algorithms provided by the authors. For the FCN-CS \cite{Elad:2017} and the proposed methods, a Nvidia GTX 1080Ti GPU is used to compute the CS reconstructions. From Table \ref{Table_time}, it can be seen that the time taken by the proposed ConvCSNet for reconstructing a $255\times 255$ image is about $1030$ times faster than the TV method \cite{LI:COA} and $760$ times faster than the D-AMP method \cite{AMP:TIF16}. Note that the high computational complexity of TV and D-AMP is not only due to slow convergence, but also the use of huge measurement matrices. When compared to the block-based FCN-CS method, our method is also about $5$ times faster.

\section{Conclusions}

Compressive sensing is a new image/signal acquisition paradigm, which has potentials in high-speed and energy efficient imaging applications. However, a practical issue in the use of CS theory is the huge memory and computation requirements for sensing a whole image. While performing CS on image block level leads to efficient measurement, it will significantly decrease the reconstruction performance. In this paper, we propose a novel convolutional compressive sensing (ConvCS) method based on deep learning. In the proposed ConvCS network, the first layer conducts sensing of the whole image using a set of convolutional filters and the remaining layers performs the reconstruction of the whole image. For better CS reconstruction, a novel two-branch convolutional neural network is proposed. Through end-to-end training, both the sensing filters and the reconstruction network can be jointly optimized. Experimental results show that the proposed method significantly outperforms existing iterative and deep learning based CS methods.

\ifCLASSOPTIONcaptionsoff
  \newpage
\fi

\bibliographystyle{IEEEtran}
\bibliography{egbib}

\end{document}